%% file: conference_101719.tex
\documentclass[conference]{IEEEtran}
\IEEEoverridecommandlockouts
\usepackage{cite}
\usepackage{amsmath,amssymb,amsfonts}
\usepackage{algorithmic}
\usepackage{graphicx}
\usepackage{textcomp}
\usepackage{enumerate}
\usepackage{booktabs}
\usepackage{multirow}
\usepackage{adjustbox}
\usepackage[dvipsnames]{xcolor}
\usepackage{amsmath, amsthm, amssymb}
\input{math_commands.tex}

\def\BibTeX{{\rm B\kern-.05em{\sc i\kern-.025em b}\kern-.08em
    T\kern-.1667em\lower.7ex\hbox{E}\kern-.125emX}}

\renewcommand{\IEEEbibitemsep}{0pt plus 0.5pt}

\makeatletter

\IEEEtriggercmd{\reset@font\normalfont\fontsize{6.5pt}{7.20pt}\selectfont}

\makeatother

\IEEEtriggeratref{1}

\begin{document}

\title{Input-Aware Dynamic Timestep Spiking Neural Networks for Efficient In-Memory Computing
\thanks{This work was supported in part by CoCoSys, a JUMP2.0 center sponsored by DARPA and SRC, Google Research Scholar Award, the NSF CAREER Award, TII (Abu Dhabi), the DARPA AI Exploration (AIE) program, and the DoE MMICC center SEA-CROGS (Award \#DE-SC0023198).}
}

\author{\IEEEauthorblockN{Yuhang Li, Abhishek Moitra, Tamar Geller, Priyadarshini Panda}
\IEEEauthorblockA{\textit{Department of Electrical Engineering, Yale University} \\
\textit{New Haven, CT 06511, USA}}
}


\maketitle

\begin{abstract}
Spiking Neural Networks (SNNs) have recently attracted widespread research interest as an efficient alternative to traditional Artificial Neural Networks (ANNs) because of their capability to process sparse and binary spike information and avoid expensive multiplication operations. 
Although the efficiency of SNNs can be realized on the In-Memory Computing (IMC) architecture, we show that the energy cost and latency of SNNs scale linearly with the number of timesteps used on IMC hardware.
Therefore, in order to maximize the efficiency of SNNs, we propose input-aware Dynamic Timestep SNN (DT-SNN), a novel algorithmic solution to dynamically determine the number of timesteps during inference on an input-dependent basis.
By calculating the entropy of the accumulated output after each timestep, we can compare it to a predefined threshold and decide if the information processed at the current timestep is sufficient for a confident prediction. We deploy DT-SNN on an IMC architecture and show that it incurs negligible computational overhead. We demonstrate that our method only uses 1.46 average timesteps to achieve the accuracy of a 4-timestep static SNN while reducing the energy-delay-product by 80\%.
\end{abstract}

\begin{IEEEkeywords}
Spiking neural networks, in-memory computing, dynamic inference
\end{IEEEkeywords}

\section{Introduction}
Deep learning has revolutionized many challenging computational tasks such as computer vision and natural language processing \cite{lecun2015deep} using Artificial Neural Networks (ANNs). These successes, however, have come at the cost of tremendous computing resources and high latency \cite{han2015deep}. Over the past decade, Spiking Neural Networks (SNNs) have gained popularity as an energy-efficient alternative to ANNs \cite{roy2019towards, tavanaei2019deep}. 
SNNs are different from ANNs in that they process inputs over a series of timesteps, whereas ANNs infer over what can be considered a single timestep.
The biologically-plausible neurons in SNNs maintain a variable called membrane potential, which controls the behavior of the SNN over a series of timesteps.
When the membrane potential exceeds a certain threshold, the neuron fires, creating a spike, and otherwise, the neuron remains inactive (neuron outputs a 0 or 1).
Such spike-based computing creates sparsity in the computations and replaces multiplications with additions.

\begin{figure}[t]
    \centering
    \includegraphics[width=\linewidth]{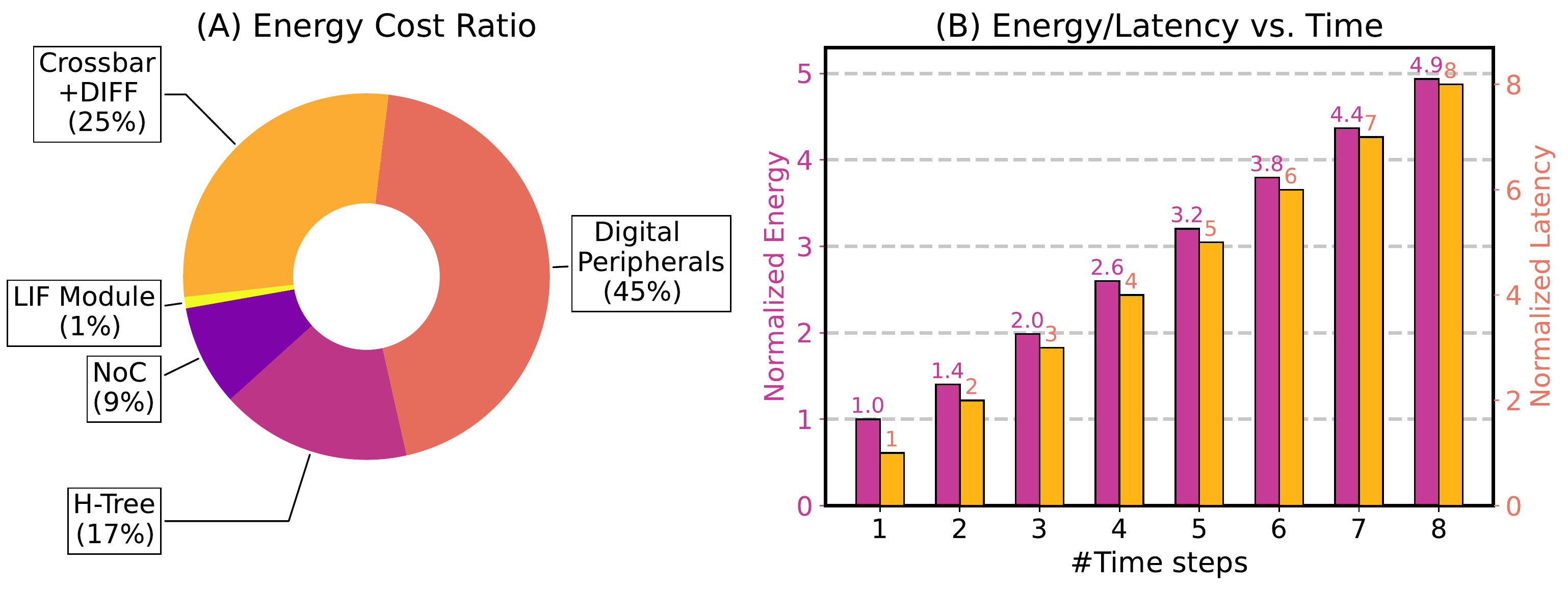}
    \caption{Energy estimation on our IMC
     architecture using VGG-16 on CIFAR-10 dataset. (A) energy ratio of each unit, (B) energy/latency vs. timesteps.}
    \label{fig_motivate}
    \vspace{-2em}
\end{figure}

Although the binary spike nature of SNNs eliminates the need for multiplications, compared to ANNs, SNNs require significantly more memory access due to multi-timestep computations on traditional von-Neumann architectures (called the “memory wall problem”) \cite{yin2022sata}. 
To alleviate this problem, In-Memory Computing (IMC) hardware is used to perform analog dot-product operations to achieve high memory bandwidth and compute parallelism \cite{sebastian2020memory}.
In this work, we mainly focus on achieving lower energy and latency in the case of IMC-implemented SNNs while maintaining iso-accuracy.

Fig.~\ref{fig_motivate}(A) shows a component-wise energy distribution for the CIFAR10-trained VGG-16 network on 64$\times$64 4-bit RRAM IMC architecture. Among these, the digital peripherals (containing crossbar input switching circuits, buffers, and accumulators) entail the highest energy cost (45\%). The IMC-crossbar and analog-to-digital converter (ADC) consumes the second highest energy (25\%). Efforts to lower energy consumption have been made by previous works. As an example, prior IMC-aware algorithm-hardware co-design techniques \cite{peng2022cmq, yuan2021tinyadc}, have used pruning, quantization to reduce the ADC and crossbar energy, and area cost. However, in the case of SNNs, the improvement is rather limited because the crossbar and ADC only occupy 25\% of the overall energy cost.

Unlike ANNs, the number of timesteps in SNNs plays an important role in hardware performance, which is orthogonal to data precision or sparsity. In Fig.~\ref{fig_motivate}(B) we investigate how timesteps affect the energy consumption and latency of an SNN. 
Note that both metrics are normalized to the performance of a 1-timestep SNN.
We find that both energy consumption and latency scale linearly with the number of timesteps, up to $4.9\times$ more energy and $8\times$ more latency when changing the number of timesteps from 1 to 8.
More importantly, if one can reduce the number of timesteps in SNNs, then all parts in Fig.~\ref{fig_motivate}(A) can benefit from the energy and latency savings.
These findings highlight the tremendous potential to optimize SNNs' performance on IMC hardware.

In fact, \cite{kim2021revisiting, li2021differentiable, chowdhury2021spatio} have explored ways to reduce the number of timesteps from an algorithmic perspective. They all train an SNN with a high number of timesteps first and then finetune the model with fewer timesteps later.
However, their method decreases the number of timesteps for all input samples, thereby inevitably leading to an accuracy-timestep trade-off. 
In this paper, we tackle this problem with another solution. 
\emph{We view the number of timesteps during inference as a variable conditional to each input sample.}
We call our method Dynamic Timestep Spiking Neural Network (DT-SNN) as it varies the number of timesteps based on each input sample. 
In particular, we use entropy thresholding to determine the appropriate number of timesteps for each sample. 
To further optimize our algorithm in practice, we design a new training loss function and implement our algorithm on an IMC architecture.

The main contributions of our work are summarized below:

\begin{enumerate}
    \item Based on what we have seen thus far, this is the first work that changes the number of timesteps in SNNs based on the input, reducing computational overhead and increasing inference efficiency without compromising task performance.
    \item To achieve that goal, we propose using entropy thresholding to distinguish the number of timesteps required. Meanwhile, we also provide a new training loss function and an IMC implementation of our method.
    \item Extensive experiments are carried out to demonstrate the efficacy and efficiency of DT-SNN. For example, the DT-SNN ResNet-19 achieves the same accuracy as the 4-timestep SNN ResNet-19 with only an average of 1.27-timestep on the CIFAR-10 dataset, reducing 84\% energy-delay-product. 
\end{enumerate}

\section{Preliminaries}

We start by introducing the basic background of SNNs. 
We denote the overall spiking neural network as a function $f_T(\vx)$ ($\vx$ is the input image), its forward propagation can be formulated as
\begin{equation}
\vy = f_T(\vx) = \frac{1}{T}\sum_{t=1}^T h\circ g^L\circ g^{L-1} \circ g^{L-2} \circ \cdots g^1(\vx),
\end{equation}
where $g^\ell(\vx) = \mathrm{LIF}(\mW^\ell \vx)$ denotes the $\ell$-th block. A block contains a convolutional layer, a leaky integrate-and-fire (LIF) layer, and an optional normalization layer placed in between the former two layers~\cite{zheng2021going}. 
$L$ represents the total number of blocks in the network and $h(\cdot)$ denotes the final linear classifier.
In this work, we use the direct encoding method, i.e., using $g^1(\vx)$ to encode the input tensor into spike trains, as done in recent SNN works~\cite{wu2019direct}. To get the final prediction, we repeat the inference process $T$ times and average the output from the classifier. 

SNNs emulate the biological neurons using LIF layers. For each timestep, the input current charges the membrane potential $\vu$ in the LIF neurons. When the membrane potential exceeds a certain threshold, a spike $\vs$ will be fired to the next layer, given by
\begin{equation}
{\vu}^\ell[t+1] = \tau\vu^\ell[t] + \mW^\ell\vs^\ell[t],
\end{equation}
\begin{equation}
\vs^{\ell+1}[t+1] = 
\begin{cases}
    1 & \text{if } \bm{u}^\ell[t+1] > V_{th} \\
    0 & \text{otherwise}
\end{cases}, \ \ 
\label{eq_fire}
\end{equation}
where $\tau\in(0, 1]$ is the leaky factor, mimicking the potential decay. If a spike is fired, the membrane potential will be reset to 0, \emph{i.e.} $(\vu[t+1]=\vu[t+1]*(1-\vs[t+1]))$. 

In the spiking neurons, all functions except the spike firing function (Eq.~(\ref{eq_fire})) can be normally differentiated. 
The firing function generates a 0 gradient if $\vu^\ell[t+1]\neq V_{th}$, otherwise, it generates a gradient with infinity. This impedes the gradient-based optimization in SNNs. 
To overcome this problem, we leverage the surrogate gradient training method \cite{wu2018spatio}. Specifically, in the forward propagation, we keep the original LIF neuron dynamics, while in the backward propagation, we use another function:
\begin{equation}
\frac{\partial \vs^\ell[t]}{\partial \vu^\ell[t]} = \max(0, V_{th} - |\vu^\ell[t]-V_{th}|)
\end{equation}



\section{Methodology}

In this section, we first introduce the algorithm for our work. Then we demonstrate the hardware implementation of DT-SNN.

\begin{figure}[t]
    \centering
    \includegraphics[width=\linewidth]{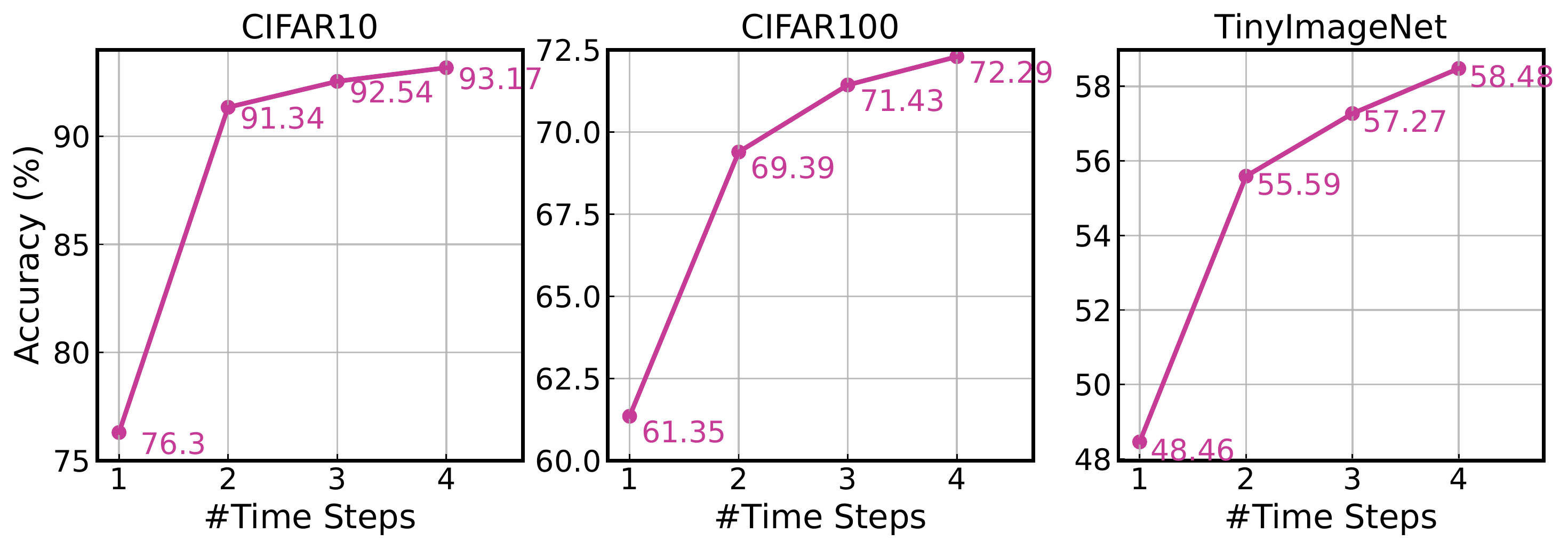}
    \caption{The impact of the number of timesteps on the accuracy. We test the spiking VGG-16 on three datasets (CIFAR10, CIFAR100, TinyImageNet).}
    \label{fig_acc_t}
\end{figure}

\subsection{Dynamic Timestep Spiking Neural Network}

Because spikes in an SNN are sparse and binary, the number of timesteps, therefore, controls the density of information inside the SNNs.
Generally, more timesteps help SNNs explore more temporal information and thus achieve higher task performance. 
Fig.~\ref{fig_acc_t} demonstrates that when the number of timesteps of an SNN VGG-16 is increased from 1 to 4 during inference, the accuracy increases as well.
Together with the hardware performance as shown in Fig.~\ref{fig_motivate}(B), the number of timesteps $T$ controls a trade-off between hardware performance and task performance on an SNN model $f_T(\cdot)$.

Unlike the conventional approach where $T$ is selected and fixed for all images, we propose a conditional probability of selecting $T$ for different input $\bm{x}$.
We call our method the Dynamic Timestep Spiking Neural Network (DT-SNN).
More concretely, denote $\mathbb{P}(T | \bm{x})$ as the conditional probability of $T$ with respect to $\bm{x}$, DT-SNN is given by
\begin{equation}
    f_{\hat{T}\sim\mathbb{P}(T | \bm{x})} = \frac{1}{\hat{T}}\sum_{t=1}^{\hat{T}} h\circ g^L\circ g^{L-1} \circ g^{L-2} \circ \cdots g^1(\vx).
\end{equation}
DT-SNN allows allocating a different number of timesteps for each input sample. 
As seen in Fig.~\ref{fig_acc_t}, we find that the majority of samples can be correctly classified with fewer timesteps. 
For example, on the CIFAR-100 dataset, 69.39\% of overall test data can be correctly predicted using only 2 timesteps.
Yet only 2.9\% of test data needs full timesteps ($T=4$) to get the right prediction. 
If we compare the hardware performance, the 4-timestep model brings 86\% more energy consumption and 100\% more latency than the 2-timestep model. 
This observation is also applicable to other datasets like CIFAR10 and TinyImageNet. 

\paragraph{Choosing the Right $T$}
Our objective in DT-SNN is to reduce the unnecessary timesteps as much as possible while not compromising accuracy. 
However, finding the appropriate timestep for each input data is non-trivial. 
In this work, we use entropy to determine $T$. 
Formally, given a dataset that has $K$ classes, the prediction probability $\pi(\bm{y}|\bm{x})$ is calculated by the Softmax function $(\sigma(\cdot))$, given by
\begin{equation}
    \pi(\bm{y}_i|\bm{x}) = \sigma_i(f(\bm{x})) = \frac{\exp(f(\bm{x})_i)}{\sum_{j=1}^K\exp(f(\bm{x})_j)},
\end{equation}
where $\pi(\bm{y}_i|\bm{x})$ is the probability of predicting $i$-th class. 
The entropy can be further calculated by
\begin{equation}
    E_f(\bm{x}) = -\frac{1}{\log K}\sum_{i=1}^K  \pi(\bm{y}_i|\bm{x})\log \pi(\bm{y}_i|\bm{x}).
    \label{eq:entropy}
\end{equation}
Here, $\log K$ ensures the final entropy is normalized to $(0, 1]$. 
The entropy measures the state of uncertainty. For instance, if all classes have an equal probability of $\frac{1}{K}$, the entropy will become 1, meaning the current state is completely random and uncertain. 
Instead, if one class's probability is approaching 1 while others are approaching 0, the entropy moves towards 0, indicating the state is becoming certain. 

Generally, the prediction accuracy is highly correlated with entropy. If the model is certain about some inputs (low entropy), the prediction would be highly probable to be correct, and vice versa \cite{guo2017calibration}. 
Therefore, we select the $T$ if the entropy is lower than some pre-defined threshold $\theta$, given by
\begin{equation}
    \hat{T}(\bm{x}) = \argmin_{\hat{T}}\{E_{f_{\hat{T}}}(\bm{x})<\theta | 1\le\hat{T} < T\} \cup \{T\}.
\end{equation}
Here, the $\hat{T}$ is selected based on the lowest timestep that can have lower-than-$\theta$ entropy. If none of them can have confident output, the SNN will use maximum timesteps, \emph{i.e.,} $T$. 

\paragraph{Training DT-SNN}
Originally, the loss function for training an SNN is the cross-entropy function, given by:
\begin{equation}
\mathcal{L}(\bm{x}, \bm{z})=-\frac{1}{B}\sum_{i=1}^K\bm{z}_i\log \pi(f_T(\bm{x})_i|\bm{x}),
\label{eq_normalloss}
\end{equation}
where $\vz$ is the label vector and $B$ is the batch size. 
Although the output from lower timesteps implicitly contributes to $f_T(\vx)$, there lacks some explicit guidance to them. 
As shown in Fig.~\ref{fig_acc_t}, the accuracy in the first timestep is always low.
Here, we propose to explicitly add a loss function to each timestep output. The new loss function is defined as:
\begin{equation}
    \mathcal{L}(\bm{x}, \bm{z}) = -\frac{1}{TB}\sum_{t=1}^T\sum_{i=1}^K \bm{z}_i\log \pi(f_t(\bm{x})_i|\bm{x}),
\label{eq_ourloss}
\end{equation}
In practice, we find this loss function does not change much training time on GPUs. Also, we will demonstrate that adding this loss function can benefit the accuracy of the outputs from all timesteps, further improving our DT-SNN. 
\paragraph{Relation to Early Exit in ANN}
Conceptually our DT-SNN is similar to the Early-Exit technique in ANN~\cite{teerapittayanon2016branchynet, bolukbasi2017adaptive} which adds multiple exits to different layers. Here, we want to clarify the relation between our DT-SNN and early exit: (1) DT-SNN is operated in the time dimension, it naturally fits SNNs and does not require any additional layers, while early exit has to add classifier layers in each branch; (2) DT-SNN has a higher potential than early exit: in experiments section, we will show that the majority of the examples can only use the first timestep, while the first exit in ANNs outputs marginal examples. 
Furthermore, DT-SNN is fully complementary to the early exit, that is, we can further add the early exit technique to SNN to fulfill even higher efficiency. 

\subsection{Hardware Implementation}
\label{sec_imc}
\begin{figure}[h!]
    \centering
    \includegraphics[width=\linewidth]{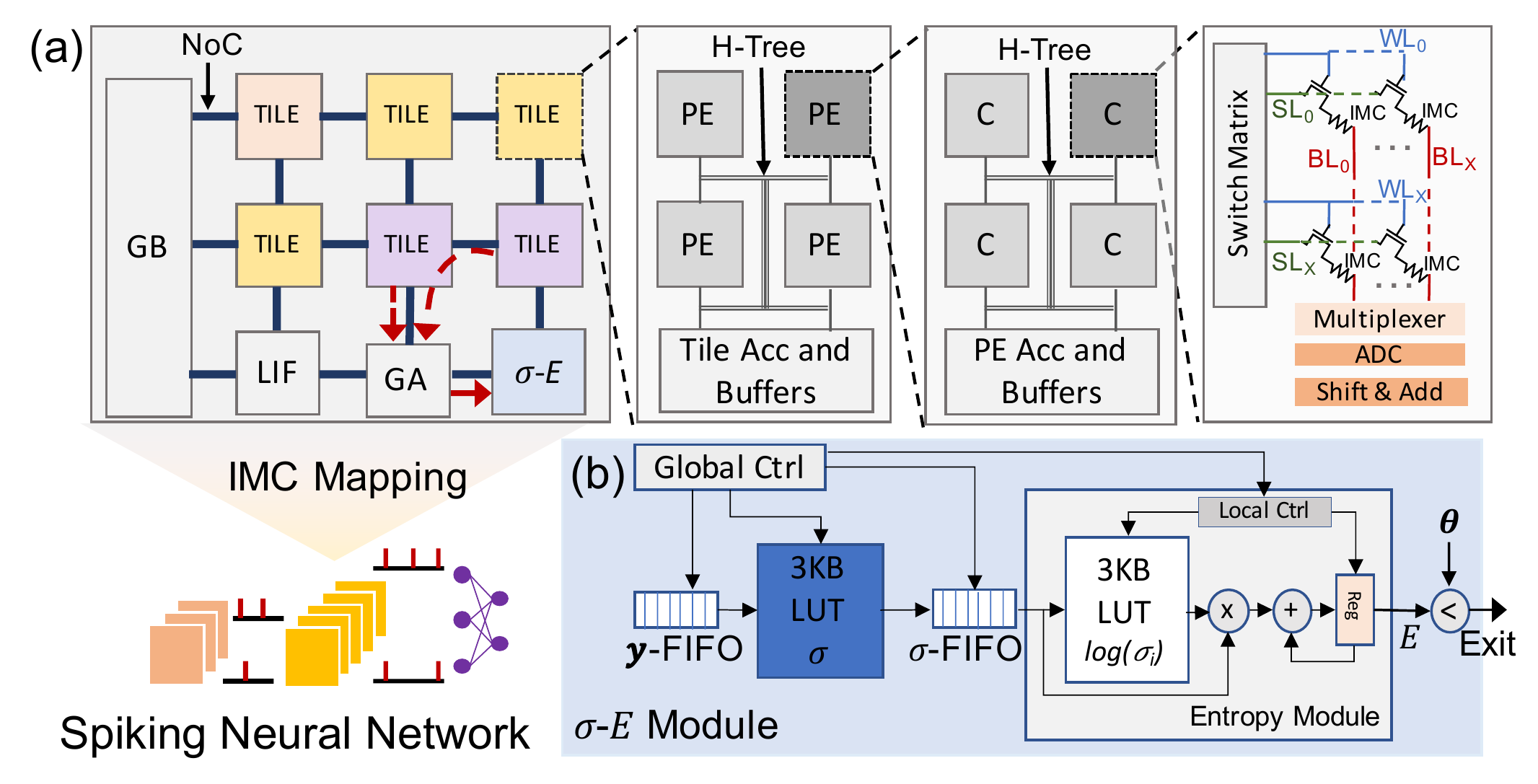}
    \caption{Figure showing (a) monolithic-tiled IMC architecture implementation of an SNN (b) architecture of the $\sigma-E$ module for softmax and entropy value computation.}
    \label{fig:hw_impl}
    \vspace{-2em}
\end{figure}

\begin{table}[h!]
    \caption{Hardware Implementation Parameters.}
    \label{tab:hw_params}
    \centering
    \begin{tabular}{|c|c|} \hline
       Technology & 32nm CMOS \\ \hline
       Crossbar Size \& Crossbars/Tile  & 64 \& 64 \\ \hline
       Device \& Weight Precision & 4-bit RRAM ($\sigma$/$\mu$=20\%) \& 8-bit \\ \hline
       $R_{off}$/$R_{on}$ & 10 at $R_{on}$=20k$\Omega$ \\ \hline
       GB, Tile \& PE Buffer Size & 20KB, 10KB \& 5KB \\ \hline
       $V_{DD}$ \& $V_{read}$ & 0.9V \& 0.1V \\ \hline
       $\sigma$ \& $E$ LUT size & 3KB \& 3KB \\ \hline
    \end{tabular}
\end{table}

\begin{table*}
\centering
\caption{Comparison between static SNN and DT-SNN in terms of timesteps, accuracy, and normalized energy cost.}
\label{tab_results}
\begin{tabular}{l l c c c c c c c c c c c c c}
\toprule
\multirow{2}{*}{\bf Model} &\multirow{2}{*}{\bf Method} &  \multicolumn{3}{c}{\bf CIFAR-10} & \multicolumn{3}{c}{\bf CIFAR-100} & \multicolumn{3}{c}{\bf TinyImageNet} & \multicolumn{3}{c}{\bf CIFAR10-DVS}\\
\cmidrule(l{2pt}r{2pt}){3-5} \cmidrule(l{2pt}r{2pt}){6-8} \cmidrule(l{2pt}r{2pt}){9-11} \cmidrule(l{2pt}r{2pt}){12-14} 
& & $T$ & Acc. & Energy & $T$ & Acc. & Energy & $T$ & Acc. & Energy & $T$ & Acc. & Energy \\
\midrule
\multirow{2}{*}{VGG-16} & SNN & 4 & 93.17 & 1.00$\times$ & 4 & 72.39 & 1.00$\times$ & 4 & 58.48 & 1.00$\times$ & 10 & 74.4 & 1.00$\times$ \\
& DT-SNN & \bf 1.46 & \bf 93.58 & \bf 0.46$\times$ & \bf 2.03 & \bf 72.43 & \bf0.56$\times$ & \bf 2.14 & \bf 58.27 & \bf 0.60$\times$ & \bf 5.25 & \bf 74.4 & \bf 0.54$\times$ \\
\midrule
\multirow{2}{*}{ResNet-19} & SNN & 4 & 94.03 & 1.00$\times$ & 4 & 73.31 & 1.00$\times$ & 4 & 56.70 & 1.00$\times$ & 10 & 75.0 & 1.00$\times$ \\
& DT-SNN & \bf 1.27 & \bf 93.87 & \bf 0.41$\times$ & \bf 1.90 & \bf 73.48 & \bf 0.53$\times$ & \bf 2.01 & \bf 56.96 & \bf 0.56$\times$ & \bf 5.02 & \bf 74.8 & \bf 0.52$\times$ \\
\bottomrule
\end{tabular}
\end{table*}

We implement DT-SNN on a tiled-monolithic chip architecture \cite{chen2018neurosim} as shown in Fig. \ref{fig:hw_impl}a. First, the individual layers of an SNN are mapped onto tiles. The number of tiles occupied by an SNN layer depends on factors such as the crossbar size, the number of input and output channels in a layer, kernel size, and the number of crossbars per tile. To implement DT-SNN-specific functionality, we incorporate the following modifications in conventionally used architectures 1) A digital $\sigma-{E}$ module to jointly compute the SoftMax ($\sigma(\cdot)$) and entropy ($E$) followed by threshold ($\theta$) comparison to detect whether to exit or not. 2) Timesteps are processed sequentially without pipelining. This eliminates the delay and hardware overhead (energy and area cost) required to empty the pipeline in case of dynamic timestep inference. The tiles additionally contain global accumulators (GA) and global buffers (GB) for accumulating partial sums and storing the intermediate outputs, respectively from different processing elements (PE). At the tile level, all modules are connected via a Network-on-Chip (NoC) interconnect. Each tile consists of several PEs, accumulators, and buffers. Each PE contains several crossbars, accumulators and buffers. The PEs and crossbars are connected by an H-Tree interconnect. We use a standard 2D-IMC crossbar connected to peripherals such as switch matrix, multiplexers, analog-to-digital converters (ADCs), and Shift-$\&$-Add circuits. The switch matrix provides input voltages at the source lines (SL) while simultaneously activating the word lines (WL). The voltages accumulate over the bit lines (BL). The analog MAC value (partial sum output) is converted to a digital value using the ADC. Multiplexers enable resource sharing of ADCs and Shift-\&-Add circuits among multiple crossbar columns to reduce the area overheads. The digital partial sum outputs from different crossbars, PEs and tiles are accumulated using the PE, tile, and global accumulators, respectively. For all layers except the last, the final MAC outputs from the GA are transferred to the LIF module for the non-linear activation functionality. The spike outputs are relayed to the tiles mapping the subsequent layers.

For the last layer (a fully connected layer) the GA accumulated MAC output is directed to the $\sigma-E$ module as shown in Fig. \ref{fig:hw_impl}a (using red arrows). Inside the $\sigma-E$ module MAC outputs are stored in the $y$-FIFO buffer. The depth of $y$-FIFO depends on the dataset. For example, in CIFAR10, the FIFO depth is 10. Data from the $y$-FIFO is passed to the address lines of the $\sigma$-LUT to compute $\sigma$ which are pushed into the $\sigma$-FIFO. The $\sigma$-FIFO outputs are sent as inputs to the Entropy Module that contains LUT for $\log(\sigma)$ computation. The Entropy Module additionally contains a multiplier and accumulator circuit (comprised of an adder and register) to implement the entropy computation using Eq.~(\ref{eq:entropy}). If the computed entropy is less than the threshold $\theta$, the inference is terminated and new input data is loaded into the GB. DT-SNN is implemented on the IMC architecture using parameters shown in Table \ref{tab:hw_params}.

\textbf{Energy Consumption of the $\sigma-E$ module:} Based on 32nm CMOS implementations, we find that the energy consumed by the $\sigma-E$ module for one timestep is merely $2e^{-5}\times$ of the 1 timestep inference energy consumed by the IMC architecture, which is negligible. 
\section{Experimental Results}

In this section, we present the evaluation of both the task performance and the hardware performance of DT-SNN, highlighting its extreme efficacy and efficiency.

\definecolor{mypink}{rgb}{0.713, 0.274, 0.5765}
\definecolor{myblue}{rgb}{0.435, 0.5765, 0.906}
\begin{figure}[t]
    \centering
    \includegraphics[width=\linewidth]{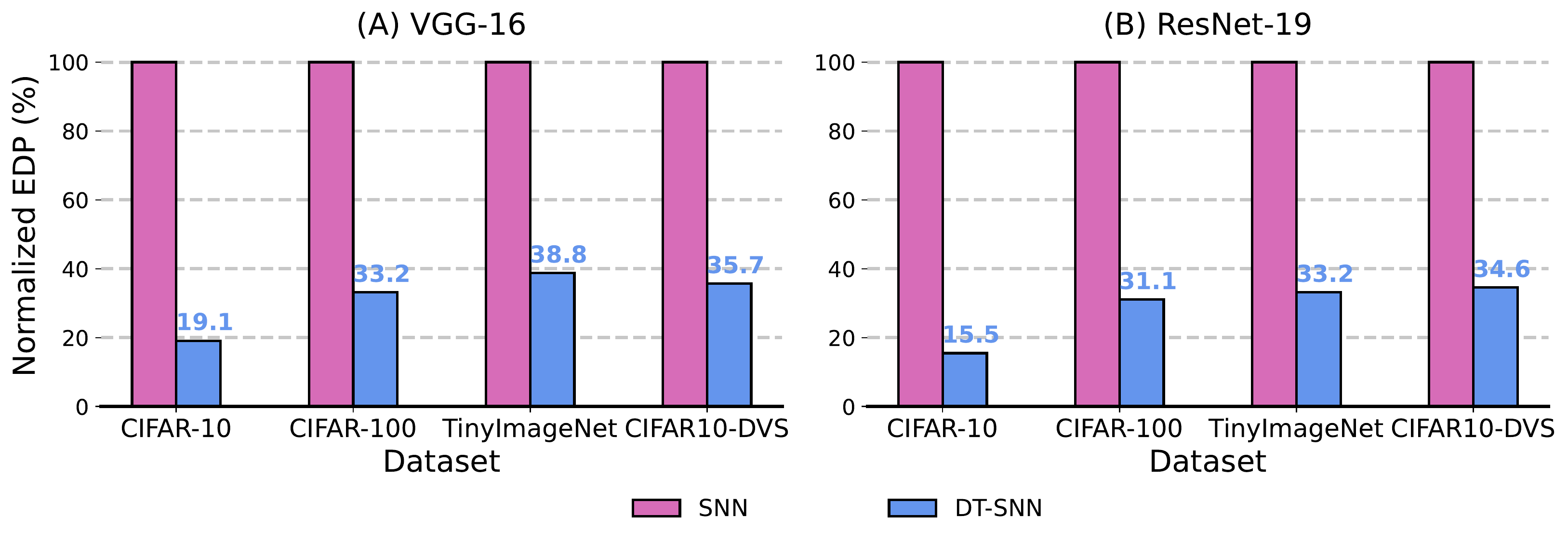}
    \caption{Comparison between static SNN and DT-SNN in terms of Energy-Delay-Product (EDP) (normalized to the static SNN). }
    \label{fig_edp_bar}
\end{figure}

\begin{figure*}
    \centering
    \includegraphics[width=0.95\textwidth]{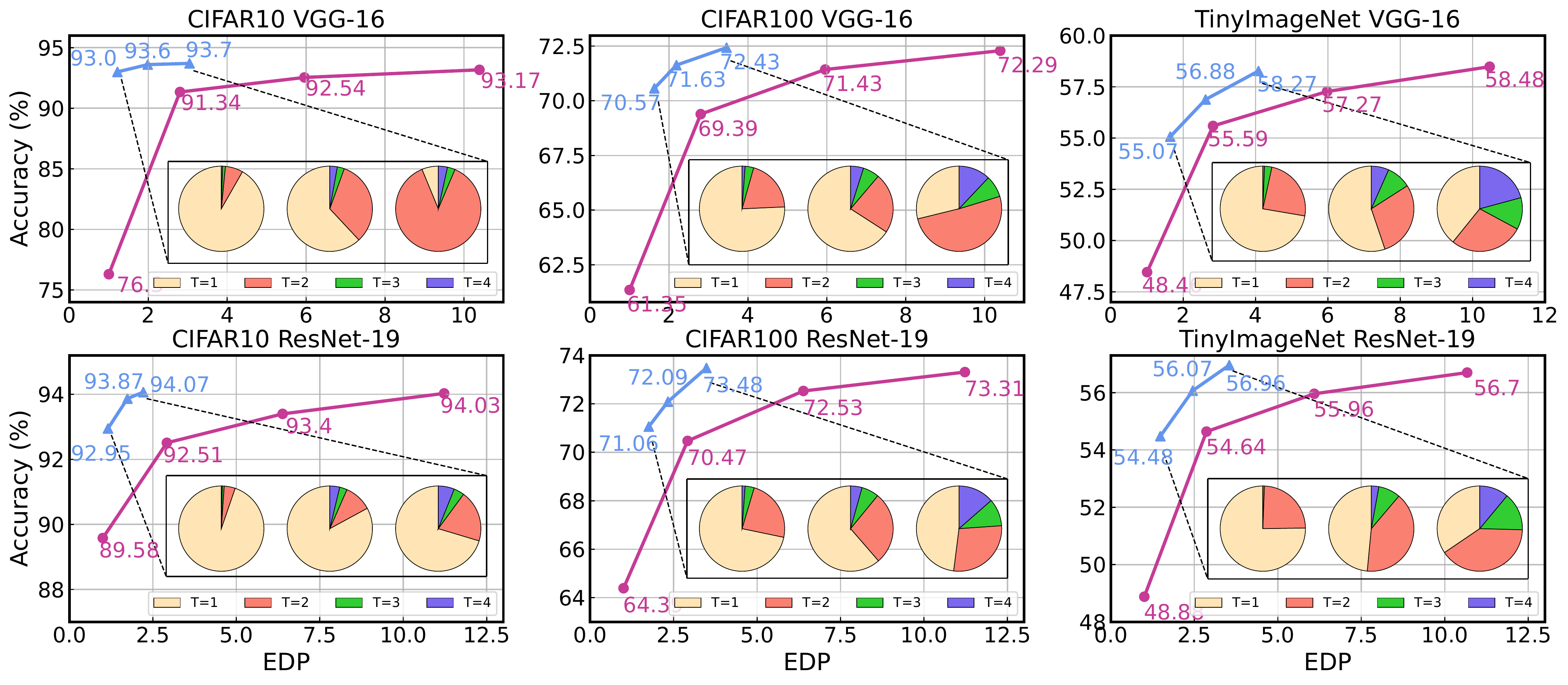}
    \vspace{-1em}
    \caption{Accuracy vs. EDP curve for both static SNN (drawn by \textcolor{mypink}{pink line}) and DT-SNN (drawn by \textcolor{myblue}{blue line}). For the static SNN, we report the accuracy and EDP when $T=\{1,2,3,4\}$. For the DT-SNN, we draw the pie charts illustrating the distribution of $\hat{T}(\vx)$.}
    \label{fig_acc_edp}
    \vspace{-1em}
\end{figure*}

\subsection{Comparison with Static SNN}

We select 4 popular visual benchmarks for evaluation, CIFAR-10~\cite{krizhevsky2009learning}, CIFAR-100~\cite{krizhevsky2009learning}, TinyImageNet~\cite{deng2009imagenet}, and CIFAR10-DVS~\cite{li2017cifar10} dataset. For the architecture, we choose to use VGG-16 \cite{simonyan2014very} and ResNet-19 \cite{he2016deep}. 
We compare our DT-SNN with the static SNN, \emph{i.e.,} an SNN that uses a fixed number of timesteps for all inputs.
The training method for the static SNN and the DT-SNN are 
kept the same, except that the static SNN uses Eq.~(\ref{eq_normalloss}) as training loss while DT-SNN uses Eq.~(\ref{eq_ourloss}). 
We train them with a batch size of 256, a learning rate of 0.1 followed by a cosine decay. The L2 regularization is set to 0.0005. 
The number of timesteps is set to 4 as done in existing work \cite{zheng2021going}. 
For task metrics, we report the top-1 accuracy on the visual benchmarks.
As for hardware metrics, we measure them based on the parameters shown in Table~\ref{tab:hw_params} and further normalize them w.r.t. static SNNs.
We report the number of timesteps ($T$), energy, and energy-delay-product (EDP).
Note that the cost of DT-SNN is varied across input data, thus we average the hardware metrics in the test dataset.

\subsubsection{Comparison of Accuracy, Energy Cost, and $T$}

We summarize the results of 4 datasets in Table \ref{tab_results}.
Here, we test static SNN with the full number of timesteps, \emph{i.e.,} $T=4$, and compare the hardware performances with DT-SNN under a similar accuracy level. 
We find that DT-SNN only needs 1.46 average timesteps on the CIFAR-10 dataset, bringing more than 50\% energy saving. 
For the other three datasets, DT-SNN requires roughly half the number of timesteps used in a static SNN model.
Nevertheless, DT-SNN reduces at least 40\% of energy cost when compared to static SNN.

\subsubsection{Comparison of EDP}
We next compare the EDP between static SNNs and DT-SNNs. 
EDP is more suitable for measuring a holistic performance in hardware because it considers both time and energy efficiency. 
Fig.~\ref{fig_edp_bar} shows the EDP comparison normalized by the EDP of the static SNN.
We can find that DT-SNN is extremely efficient as it reduces 61.2\%$\sim$80.9\% EDP of static SNNs.
These results highlight the efficiency brought by our method, leading to both energy cost and latency reduction. 

\subsubsection{Accuracy vs. EDP curve}

The static SNN can adjust the number of timesteps for all inputs to balance its accuracy and efficiency.
Our DT-SNN can also adjust threshold $\theta$ to balance such a trade-off.
Here, we draw an accuracy-EDP curve in Fig.~\ref{fig_acc_edp}. 
\emph{Note that here the EDP is normalized to the EDP of the 1-timestep static SNN.} 
We evaluate static SNN at 1,2,3, and 4 timesteps, and evaluate DT-SNN using three different thresholds. 
It can be seen that our DT-SNN is placed in the top-left corner, indicating a better accuracy-EDP trade-off than the static SNN. 
Remarkably, DT-SNN can bring significant improvement in low-timestep scenarios. 
For instance, DT-SNN VGG-16 increases the accuracy by 17\% when compared to the 1-timestep static counterpart on the CIFAR-10 dataset, while it only has $\sim$10\% higher EDP.

In order to further visualize the dynamic timesteps in our method, Fig.~\ref{fig_acc_edp} provides three pie charts in each case, which show the percentage of input examples that are inferred with 1, 2, 3, or 4 timesteps. 
Notably, $T=1$ is usually the most selected case for the input due to the fact that most input examples can be correctly predicted using only 1 timestep.
As the threshold decreases, more images start to use higher timesteps. 
Overall, we find $T=3$ and $T=4$ are rarely used in DT-SNN, demonstrating the effectiveness of our method to reduce redundant timesteps.

\begin{figure}[t]
    \centering
    \includegraphics[width=\linewidth]{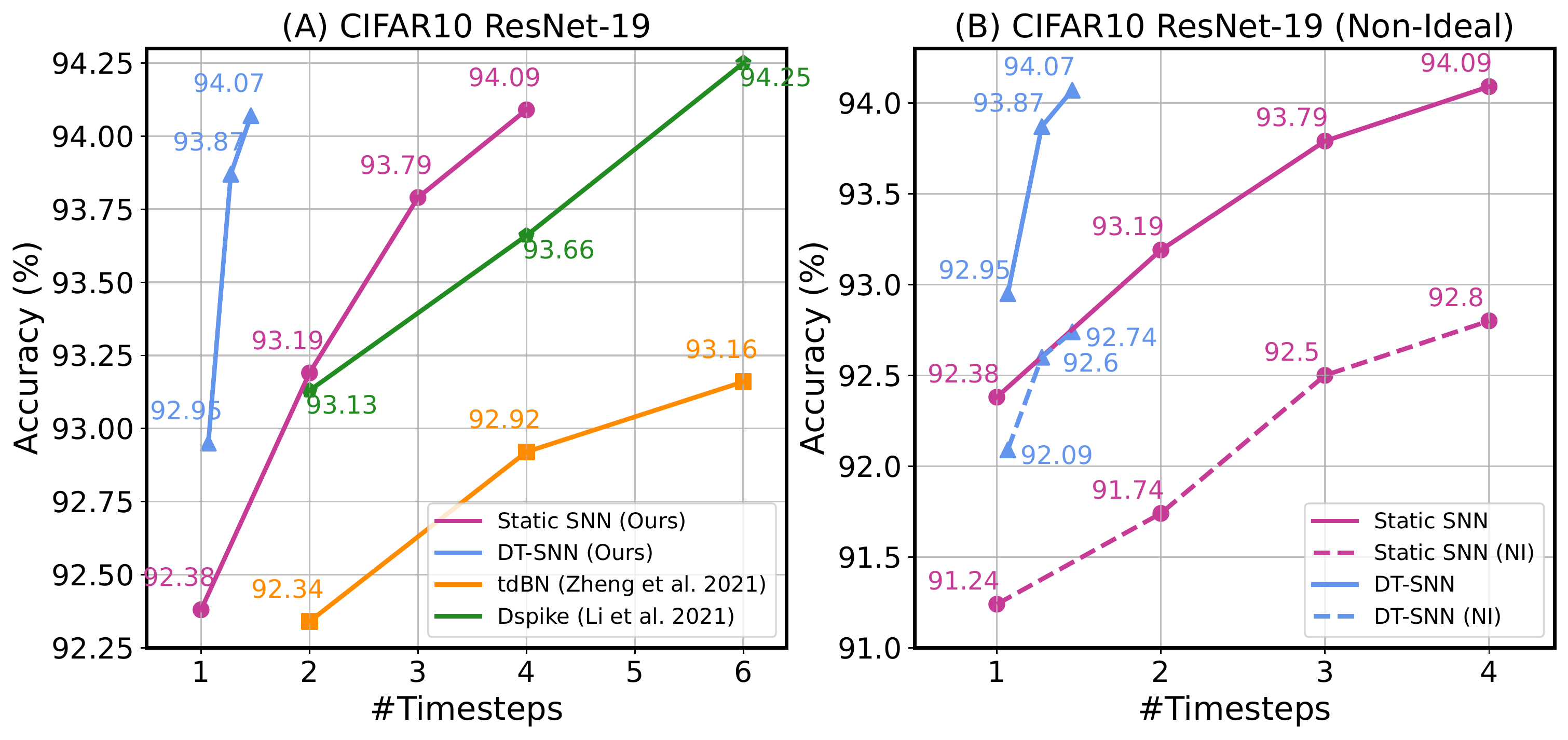}
    \caption{Accuracy vs. the number of timesteps. (A) Comparison with prior works, (B) Comparison under non-ideal (NI) device variation in the IMC.}
    \label{fig_acc_tab}
    \vspace{-1em}
\end{figure}

\subsubsection{Comparison with Prior Work }
Here, we also compare our method with prior work on SNNs. We compare tdBN~\cite{zheng2021going} and Dspike~\cite{li2021differentiable} with our static-SNN and DT-SNN trained with Eq.~(\ref{eq_ourloss}). 
Fig.~\ref{fig_acc_tab}(A) shows the accuracy under different $T$. Our DT-SNN reaches a new state of the art. 

\subsubsection{Non-Ideal Accuracy with Device Variations}
So far, all the experiments are performed without considering device conductance variation. In Fig.~\ref{fig_acc_tab}(B), we compare DT-SNN and static SNN under 20\% device conductance variation.
We simulate this by adding noise to the weights post-training. We see that DT-SNN still maintains higher accuracy while eliminating the redundant timesteps compared to static-SNN .

\subsection{Acceleration in General Processors}

\begin{table}[t]
\centering
\caption{Throughput comparison on the GPU.}
\adjustbox{max width=\linewidth}{
\begin{tabular}{l|ccc|ccc}
\toprule
\bf Dataset\&Model & \multicolumn{3}{c|}{\bf CIFAR10 VGG-16} & \multicolumn{3}{c}{\bf CIFAR10 ResNet-19}\\
\midrule
\bf Method & $T$ & Acc. & Throughput & $T$ & Acc. & Throughput \\
\midrule
\multirow{4}{*}{SNN} & 1 & 76.30 & 199.3 & 1 & 89.58 & 185.3 \\
& 2 & 91.34 & 121.8 & 2 & 92.51 & 105.9\\
& 3 & 92.54 & 85.19 & 3 & 93.40 & 77.14 \\
& 4 & 93.01 & 64.34 & 4 & 94.03 & 62.65\\
\midrule
\multirow{3}{*}{DT-SNN} & 1.10 & 93.01 & 176.7 & 1.07 & 92.95 & 169.3 \\
& 1.46 & 93.58 & 142.0 & 1.27 & 93.87 & 147.5\\
& 2.11 & 93.71 & 105.9 & 1.46 & 94.07 & 135.7\\
\bottomrule
\end{tabular}}
\label{tab_throughput}
\end{table}

In the previous section, we demonstrated that our DT-SNN can accelerate the inference on an IMC architecture. 
Here, we show that apart from the IMC architecture we simulated, our method can also be applicable to other types of hardware like digital processors. 
To this end, we measure the inference throughput (\emph{i.e.,} the number of images inferred per second) on an RTX 2080Ti GPU simulated by PyTorch using batch size 1. 
Table \ref{tab_throughput} lists the accuracy, (averaged) timesteps, and throughput. 
As can be seen from the table, the throughput on GPU significantly reduces as the timesteps are increased.
Compared to static SNNs, our DT-SNN substantially improves the throughput while not sacrificing accuracy. 
For example, DT-SNN ResNet-19 with 1.07 averaged timesteps can infer 169.3 images per second, quite close to the 1-timestep static SNN (185.3 images per second), yet still bringing 3.4\% accuracy improvement. 

\subsection{Ablation Study}

\begin{figure}[t]
    \centering
    \includegraphics[width=\linewidth]{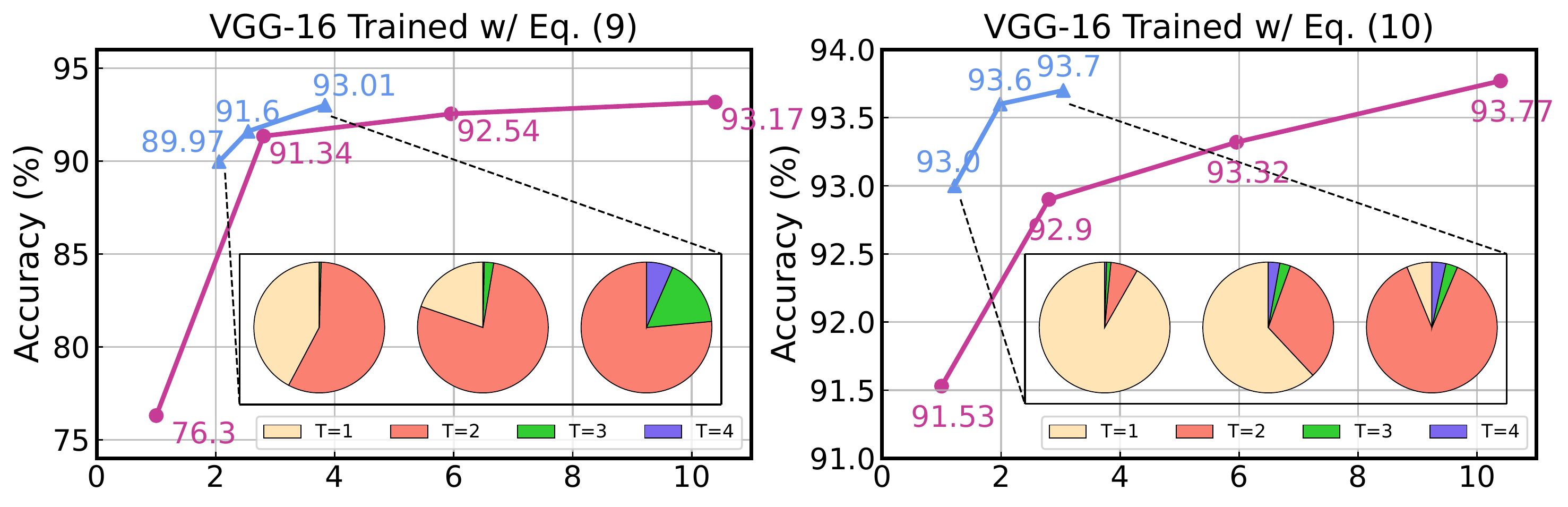}
    \caption{Comparison between two loss functions using accuracy-EDP curves on the CIFAR-10 dataset. Annotations are aligned with Fig.~\ref{fig_acc_edp}.}
    \label{fig_ablation}
\end{figure}

In this section, we ablate the training loss function choice. To verify this, we train static an SNN VGG-16 on CIFAR-10 either with Eq.~(\ref{eq_normalloss}) or Eq.~(\ref{eq_ourloss}) and test the corresponding accuracies of both static SNN and DT-SNN. 
Fig.~\ref{fig_ablation} demonstrates the comparison between these two loss functions.
We find that our training loss boosts the accuracy of all timesteps in the static SNN. 
In particular, the first timestep of VGG-16 changes from 76.3\% to 91.5\%. This result proves that explicit guidance from label annotations should be added to the lower timesteps.
Meanwhile, it also increases the full timestep performance, resulting in a 0.6\% accuracy increase on VGG-16.

This improvement is extremely beneficial to DT-SNN. 
According to the pie charts describing the distribution of $\hat{T}$, we find our training loss function enables a smaller number of timesteps to be required to classify the test data, thus reducing the EDP significantly.


\subsection{Visualization}

\begin{figure}[t]
    \centering
    \includegraphics[width=0.48\linewidth]{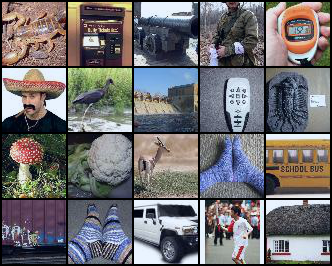}
    \hfill
    \includegraphics[width=0.48\linewidth]{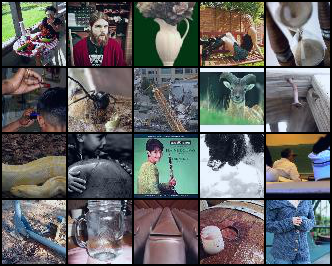}
    \caption{Visualization of input samples from $T=1$ (left) and $T=4$ (right).}
    \label{fig_visualization}
\end{figure}

In this section, we visualize the input images that are differentiated by our DT-SNN. 
Ideally, we anticipate DT-SNN can identify whether an image is easy or hard to infer (corresponding to 4 or 1 timestep). 
To maximize the differentiation, we use a low threshold to filter out the high timesteps, so that only the easiest images can be classified in the first timestep and vice versa.
Fig. \ref{fig_visualization} presents the results on the TinyImageNet dataset. 
Generally, we find the images inferred with 1 timestep exhibit a simple structure: a clear object placed in the center of a clean background. 
In contrast, hard images require 4 timesteps and usually mix the background and the object together, making the object imperceptible. 

\section{Conclusion}
In this work, we introduce the Dynamic Timestep Spiking Neural Network, a simple yet significantly effective method that selects different timesteps for different input samples. 
DT-SNN determines the suitable timestep based on the confidence of the model output, seamlessly fitting the nature of sequential processing over time in SNNs. 
Moreover, DT-SNN is practical. It can be deployed to IMC architecture and even general digital processors.
Extensive experiments prove that DT-SNN establishes a new state-of-the-art trade-off between hardware efficiency and task performance.

{
\bibliographystyle{plain} 
\bibliography{references} 
}


\end{document}

%% file: math_commands.tex
\usepackage{amsmath,amsfonts,bm}

\def\eqref#1{equation~\ref{#1}}

\def\1{\bm{1}}

\def\vs{{\bm{s}}}

\def\vu{{\bm{u}}}

\def\vx{{\bm{x}}}
\def\vy{{\bm{y}}}
\def\vz{{\bm{z}}}

\def\mW{{\bm{W}}}

\DeclareMathAlphabet{\mathsfit}{\encodingdefault}{\sfdefault}{m}{sl}
\SetMathAlphabet{\mathsfit}{bold}{\encodingdefault}{\sfdefault}{bx}{n}

\DeclareMathOperator*{\argmin}{arg\,min}

%% file: conference_101719.bbl
\begin{thebibliography}{10}

\bibitem{bolukbasi2017adaptive}
Tolga Bolukbasi et~al.
\newblock Adaptive neural networks for efficient inference.
\newblock In {\em ICML}, pages 527--536. PMLR, 2017.

\bibitem{chen2018neurosim}
Pai-Yu Chen et~al.
\newblock Neurosim: A circuit-level macro model for benchmarking neuro-inspired
  architectures in online learning.
\newblock {\em IEEE TCAD}, 37(12):3067--3080, 2018.

\bibitem{chowdhury2021spatio}
Sayeed~Shafayet Chowdhury et~al.
\newblock Spatio-temporal pruning and quantization for low-latency spiking
  neural networks.
\newblock In {\em 2021 IJCNN}, pages 1--9. IEEE, 2021.

\bibitem{deng2009imagenet}
Jia Deng et~al.
\newblock Imagenet: A large-scale hierarchical image database.
\newblock In {\em 2009 CVPR}, pages 248--255. Ieee, 2009.

\bibitem{guo2017calibration}
Chuan Guo et~al.
\newblock On calibration of modern neural networks.
\newblock In {\em ICML}, pages 1321--1330. PMLR, 2017.

\bibitem{han2015deep}
Song Han et~al.
\newblock Deep compression: Compressing deep neural networks with pruning,
  trained quantization and huffman coding.
\newblock {\em arXiv preprint arXiv:1510.00149}, 2015.

\bibitem{he2016deep}
Kaiming He et~al.
\newblock Deep residual learning for image recognition.
\newblock In {\em Proceedings of the IEEE conference on CVPR}, pages 770--778,
  2016.

\bibitem{kim2021revisiting}
Youngeun Kim et~al.
\newblock Revisiting batch normalization for training low-latency deep spiking
  neural networks from scratch.
\newblock {\em Frontiers in neuroscience}, page 1638, 2021.

\bibitem{krizhevsky2009learning}
Alex Krizhevsky, Geoffrey Hinton, et~al.
\newblock Learning multiple layers of features from tiny images.
\newblock 2009.

\bibitem{lecun2015deep}
Yann LeCun et~al.
\newblock Deep learning.
\newblock {\em nature}, 521(7553):436--444, 2015.

\bibitem{li2017cifar10}
Hongmin Li et~al.
\newblock Cifar10-dvs: an event-stream dataset for object classification.
\newblock {\em Frontiers in neuroscience}, 11:309, 2017.

\bibitem{li2021differentiable}
Yuhang Li et~al.
\newblock Differentiable spike: Rethinking gradient-descent for training
  spiking neural networks.
\newblock {\em NeurIPS}, 34, 2021.

\bibitem{peng2022cmq}
Jie Peng et~al.
\newblock Cmq: Crossbar-aware neural network mixed-precision quantization via
  differentiable architecture search.
\newblock {\em IEEE TCAD}, 2022.

\bibitem{roy2019towards}
Kaushik Roy et~al.
\newblock Towards spike-based machine intelligence with neuromorphic computing.
\newblock {\em Nature}, 575(7784):607--617, 2019.

\bibitem{sebastian2020memory}
Abu Sebastian et~al.
\newblock Memory devices and applications for in-memory computing.
\newblock {\em Nature nanotechnology}, 15(7):529--544, 2020.

\bibitem{simonyan2014very}
Karen Simonyan et~al.
\newblock Very deep convolutional networks for large-scale image recognition.
\newblock {\em arXiv preprint arXiv:1409.1556}, 2014.

\bibitem{tavanaei2019deep}
Amirhossein Tavanaei et~al.
\newblock Deep learning in spiking neural networks.
\newblock {\em Neural Networks}, 111:47--63, 2019.

\bibitem{teerapittayanon2016branchynet}
Surat Teerapittayanon et~al.
\newblock Branchynet: Fast inference via early exiting from deep neural
  networks.
\newblock In {\em 2016 23rd ICPR}, pages 2464--2469. IEEE, 2016.

\bibitem{wu2018spatio}
Yujie Wu et~al.
\newblock Spatio-temporal backpropagation for training high-performance spiking
  neural networks.
\newblock {\em Frontiers in neuroscience}, 12:331, 2018.

\bibitem{wu2019direct}
Yujie Wu et~al.
\newblock Direct training for spiking neural networks: Faster, larger, better.
\newblock In {\em AAAI}, volume~33, pages 1311--1318, 2019.

\bibitem{yin2022sata}
Ruokai Yin et~al.
\newblock Sata: Sparsity-aware training accelerator for spiking neural
  networks.
\newblock {\em arXiv preprint arXiv:2204.05422}, 2022.

\bibitem{yuan2021tinyadc}
Geng Yuan et~al.
\newblock Tinyadc: Peripheral circuit-aware weight pruning framework for
  mixed-signal dnn accelerators.
\newblock In {\em 2021 DATE}, pages 926--931. IEEE, 2021.

\bibitem{zheng2021going}
Hanle Zheng et~al.
\newblock Going deeper with directly-trained larger spiking neural networks.
\newblock In {\em AAAI}, volume~35, pages 11062--11070, 2021.

\end{thebibliography}
